\documentclass[10pt,twocolumn,letterpaper]{article}

\usepackage{iccv}              
\usepackage[most]{tcolorbox}
\usepackage{multirow}
\usepackage{colortbl}
\usepackage{xcolor}
\usepackage{array} 
\usepackage{colortbl} 
\usepackage{booktabs}
\usepackage{tabularx}

\definecolor{iccvblue}{rgb}{0.21,0.49,0.74}
\usepackage[pagebackref,breaklinks,colorlinks,allcolors=iccvblue]{hyperref}

\def\paperID{10507} 
\def\confName{ICCV}
\def\confYear{2025}

\title{Bridging the Gap Between Ideal and Real-world Evaluation: Benchmarking AI-Generated Image Detection in Challenging Scenarios}

\author{
Chunxiao Li\textsuperscript{1}\thanks{Equal contribution.} \quad
Xiaoxiao Wang\textsuperscript{2,*} \quad
Meiling Li\textsuperscript{3} \quad
Boming Miao\textsuperscript{1} \\
Peng Sun\textsuperscript{4} \quad
Yunjian Zhang\textsuperscript{5} \quad
Xiangyang Ji\textsuperscript{5} \quad
Yao Zhu\textsuperscript{5}\thanks{Corresponding author.}
\\[1.5ex]
\textsuperscript{1}Beijing Normal University, Beijing, China \quad
\textsuperscript{2}University of Chinese Academy of Sciences, Beijing, China \\
\textsuperscript{3}Fudan University, Shanghai, China \quad
\textsuperscript{4}Central University of Finance and Economics, Beijing, China \\
\textsuperscript{5}Tsinghua University, Beijing, China \\
{\tt\small chunxiaoli@mail.bnu.edu.cn, ee$\_$zhuy@zju.edu.cn
}}

\begin{document}
\maketitle
\begin{abstract}
With the rapid advancement of generative models, highly realistic image synthesis has posed new challenges to digital security and media credibility. Although AI-generated image detection methods have partially addressed these concerns, a substantial research gap remains in evaluating their performance under complex real-world conditions. This paper introduces the Real-World Robustness Dataset (RRDataset) for comprehensive evaluation of detection models across three dimensions: 1) \textbf{Scenario Generalization} – RRDataset encompasses high-quality images from seven major scenarios (War \& Conflict, Disasters \& Accidents, Political \& Social Events, Medical \& Public Health, Culture \& Religion, Labor \& Production, and everyday life), addressing existing dataset gaps from a content perspective. 2) \textbf{Internet Transmission Robustness} – examining detector performance on images that have undergone multiple rounds of sharing across various social media platforms.
3) \textbf{Re-digitization Robustness} – assessing model effectiveness on images altered through four distinct re-digitization methods.

We benchmarked 17 detectors and 10 vision-language models (VLMs) on RRDataset and conducted a large-scale human study involving 192 participants to investigate human few-shot learning capabilities in detecting AI-generated images. The benchmarking results reveal the limitations of current AI detection methods under real-world conditions and underscore the importance of drawing on human adaptability to develop more robust detection algorithms. Our dataset is publicly available at:  \url{https://zenodo.org/records/14963880}.

\end{abstract}    
\section{Introduction}
\label{sec:intro}
\begin{figure*}[htbp]
  \centering
  \includegraphics[width=0.9\textwidth]{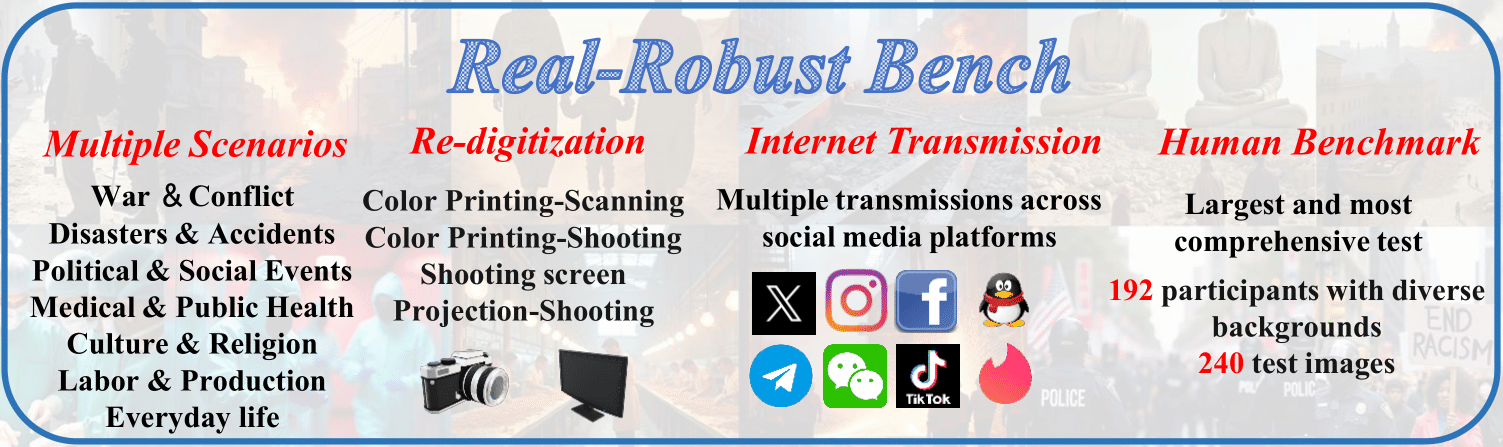}
  \caption{Real-Robust Bench overview, featuring a wide range of scenarios and real-world robustness incorporating four re-digitization methods and internet transmission processes. Additionally, we introduce the largest human benchmark to date, with 192 participants from diverse backgrounds and 240 test images.}
  \label{fig:intro}
\end{figure*}

With the emergence of powerful generative models \cite{rombach2022high, karras2022elucidating, peebles2023scalable,saharia2022photorealistic}, AI-generated images are increasingly difficult to distinguish from real ones. Such indistinguishable AI-generated images pose risks to society, including the spread of misinformation and misleading visual cues. Consequently, the identification of AI-generated images has become a critical task with profound real-world implications.

An increasing number of studies explore various detection methods based on different features, including image texture \cite{liu2020global}, gradients \cite{tan2023learning}, patch-level features \cite{chai2020makes, mandelli2022detecting, chen2024single}, frequency-domain characteristics \cite{zhang2019detecting, durall2020watch, dzanic2020fourier, frank2020leveraging}, fine-grained image details \cite{sarkar2024shadows, ju2022fusing}, and reconstruction loss \cite{wang2023dire, ricker2024aeroblade}. Vision-language models (VLMs) have also been adapted for detection tasks \cite{khan2024clipping, keita2024harnessing, sha2023fake, ojha2023towards}. Although conventional benchmarks such as GenImage \cite{zhu2024genimage}, Fake2M \cite{lu2024seeing}, WildFake \cite{hong2024wildfake}, and Chameleon \cite{yan2024sanity} offer sizeable collections of generative images, they share two significant limitations. First, they primarily consist of everyday-life images, making it hard to assess whether the detection methods can effectively generalize to other scenarios. Second, they fail to account for the influence of internet social media transmissions or re-digitization processes, thus inflating detection performance relative to the complexities of real-world conditions.

To address these gaps, we introduce RRDataset, the first benchmark designed explicitly to evaluate detectors’ robustness in practical contexts. RRDataset encompasses images from seven diverse and challenging scenarios—including  War \& Conflict, Disasters \& Accidents, Political \& Social Events, Medical \& Public Health, Culture \& Religion, and Labor \& Production—thereby filling crucial content gaps in existing benchmarks. Moreover, it systematically includes internet transmission and four distinct re-digitization processes, enabling a deeper assessment of how these factors degrade detector performance in realistic use cases. We benchmark 17 detection methods and 10 VLMs on RRDataset, finding substantial performance declines under both transmission and re-digitization conditions—shortcomings that have largely been overlooked in previous evaluations. We further establish a large-scale human benchmark, involving 192 participants and 240 test images, to investigate how human observers adapt in similar conditions. Although human accuracy also decreases when confronted with altered images, it improves significantly after a few-shot learning phase, highlighting a capacity for rapid adaptation that may inspire future detection algorithm design. The main contributions are summarized as follows: 
\begin{itemize} 
    \item We propose RRDataset, the first dataset incorporating six critical high-risk scenarios and aspects of daily life. To enhance its realism and applicability, the images within this dataset have been transmitted over the internet and re-digitized, thereby simulating real-world image degradation as shown in Fig. \ref{fig:intro}. 
    \item We conduct RRBench, a comprehensive benchmark evaluating 17 detection methods and 10 VLMs, revealing that current detection strategies suffer notable performance drops under transmission and re-digitization. 
    \item Our RRBench also constructs the largest human AI-Generated image benchmark to date, involving 192 participants and 240 test images. Our analysis uncovers the remarkable few-shot adaptation capabilities of human observers.
\end{itemize}

\section{Related work}

\begin{figure*}[!htbp]
  \centering
  \includegraphics[width=1.0\textwidth]{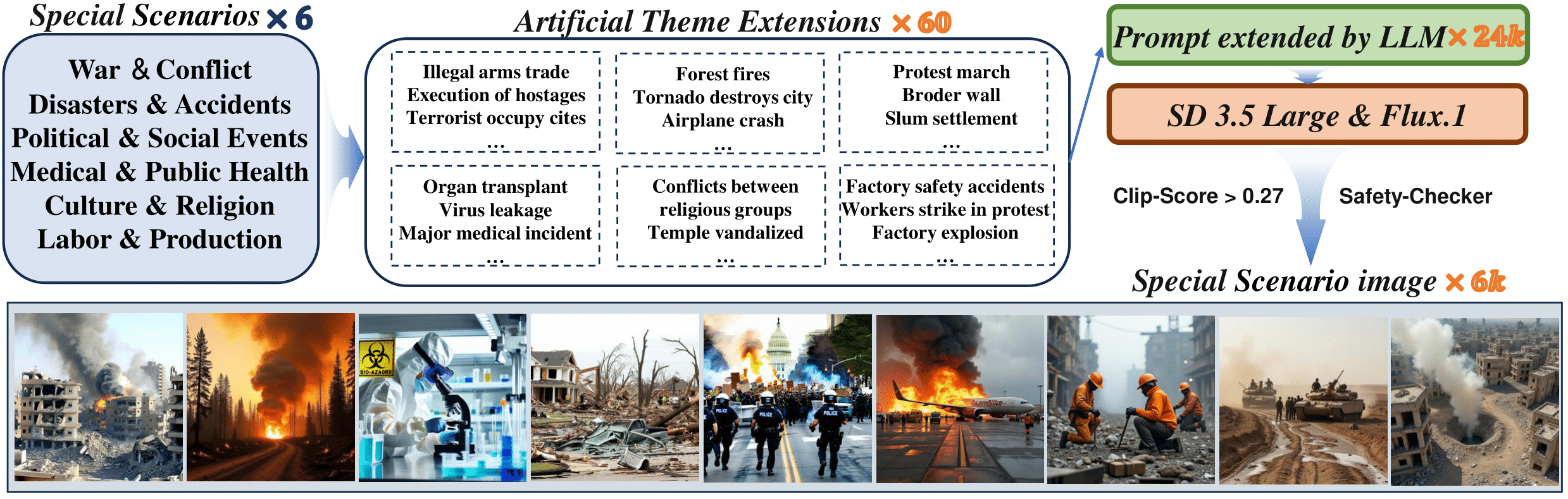}
  \caption{Special-scenario image generation pipeline, illustrating scenario definition, theme expansion, prompt refinement, and final image filtering.}
  \label{fig:datapipe}
\end{figure*}

\textbf{AI-Generated Image Datasets:} With the rapid evolution of generative models, AI-generated image detection datasets have undergone substantial changes. Early efforts, such as CNNSpot \cite{wang2020cnn}, primarily employed ProGAN \cite{karras2017progressive}-produced images to evaluate detection performance. More recent works—including CiFAKE \cite{bird2024cifake} and DE-FAKE \cite{sha2023fake}—have shifted toward diffusion-based approaches, yielding more realistic synthetic images.

Large-scale datasets such as GenImage \cite{zhu2024genimage}, WildFake \cite{hong2024wildfake}, Fake2M \cite{lu2024seeing}, Chameleon \cite{yan2024sanity}, and PatchCraft \cite{zhong2024patchcraft} have pushed the boundaries of size and diversity, reaching millions of synthetic samples from multiple model architectures. Some datasets like Chameleon \cite{yan2024sanity}, emphasize high-quality human-curated images to expose blind spots in current detectors. However, most of these datasets concentrate on everyday scenarios drawn from sources like ImageNet \cite{deng2009imagenet} or COCO \cite{lin2014microsoft}, neglecting critical contexts such as war, disasters, or public health, where misinformation can be particularly damaging. 

Several recent works also explore robustness under data augmentation or online transmission. SEMI-TRUTHS \cite{pal2025semi} investigates how different augmentations affect detection performance, whereas WildRF \cite{cavia2024real} and FOSID \cite{karageorgiou2024evolution} collect a limited amount of web-transmitted data to examine detector robustness. However, their choice of platforms and overall scale remains narrow, rendering the results less representative of the broader impact of social media transmission.

RRDataset addresses these gaps by covering diverse high-impact scenarios while also modeling multiple rounds of online transmission and various re-digitization processes. This approach enables a more faithful assessment of detection performance in scenarios that closely mirror actual practices and challenges.

\noindent \textbf{Human Benchmark for AI Image Detection:} HPBench \cite{lu2024seeing} is the first to collect data from 50 participants on 100 test images, along with insights into why participants identified certain images as AI-generated. However, this study used images exclusively generated by Midjourney, which may introduce bias due to reliance on a single generator, and its limited sample size and question set constrain the scope of its findings. While FakeBench \cite{li2024fakebench} uses images from multiple generators, it does not account for the effects of image transmission or re-digitization and includes only 34 participants. This paper establish the largest human benchmark to date, addressing these limitations by incorporating 192 participants and 240 test images, encompassing original, transmitted, and re-digitized images. We conducted a detailed analysis of how AIGC and photography backgrounds impact detection accuracy. Additionally, we explored human few-shot learning capabilities in detecting AI-generated images.

\noindent \textbf{AI-Generated Image Detectors:} With the continuous development of generative models, numerous detection methods \cite{liu2020global,chai2020makes,mandelli2022detecting,ju2022fusing,corvi2023detection,ojha2023towards,wang2023dire,chen2024single,sarkar2024shadows,ricker2024aeroblade,tan2024rethinking,chen2024drct,koutlis2024leveraging,tan2024c2p,yan2024sanity,li2024improving} have emerged. While these methods achieve impressive accuracy on their respective test sets, robust evaluation remains largely unaddressed. Existing methods typically limit robustness evaluation to resizing, JPEG compression and Gaussian blur, with some studies using only JPEG compression at a quality level of 90, which does not closely simulate real-world conditions. Therefore, it is essential to evaluate these detectors’ performance in realistic scenarios. A more comprehensive overview of AI-generated image detection methods can be found in App. A.

\section{Dataset Construction}
Sec. \ref{sec31} describes the problem setting and design rationale. Sec. \ref{sec32} outlines our collection and creation process, and Sec. \ref{sec33} explains internet transmission and re-digitization. Additional examples are included in App. G.

\subsection{Dataset Rationale}
\label{sec31}
The primary goal of RRDataset is to comprehensively benchmark AI-generated image detectors under real-world conditions, focusing on three critical dimensions: 1) \textbf{Multi-scenario detection capability: } We incorporate a broad range of image contexts, including rarely encountered or sensitive scenarios, to ensure that detection models are evaluated on content closely mirroring real-world usage. 2) \textbf{Internet transmission robustness: } Images routinely undergo repeated sharing and compression on various social media platforms. We expect a reliable detector to maintain accuracy despite these transmission-induced degradations. 3) \textbf{Re-digitization robustness: } Re-digitization—via scanning, re-photographing, or similar methods—is a pivotal yet underexplored challenge. Many practical scenarios lack direct digital files (e.g., PNG, JPG), such as verifying images in newspapers, presentation slides, or street advertisements. In RRDataset, an image’s label (Real vs. AI) is determined solely by its original source, underscoring the need for detectors to withstand transformations introduced by re-digitization.

By encompassing these three dimensions, RRDataset aims to provide a comprehensive and realistic evaluation framework that addresses the often-overlooked factors limiting current detection methods in existing assessments.

\subsection{Data Collection}
\label{sec32}
\subsubsection{Special-Scenario Image Collection}
To capture diverse and critical contexts, RRDataset includes six specialized scenarios: War \& Conflict, Disasters \& Accidents, Political \& Social Events, Medical \& Public Health, Culture \& Religion, and Labor \& Production. First, each scenario is expanded into 10 manually defined theme (e.g., traffic accidents, floods, earthquakes, wildfires, and plane crashes under Disasters \& Accidents).
Next, we use Qwen2.57b-instruct \cite{yang2024qwen2} to further enrich these sub-topics into prompts, adding location details and additional descriptive elements. This process yields 24,000 prompts, which are then used by SD 3.5 Large \cite{rombach2022high} and Flux.1 \cite{flux2024} to generate 48,000 images at various resolutions.
We filter the generated images using CLIP-score \cite{radford2021learning} (removing those below 0.27) and an NSFW safety checker, discarding any that exhibit low text-image alignment or contain unsafe content. After filtering, we retain 6,000 high-quality AI-generated images (1,000 per scenario), the complete image generation workflow is illustrated in Fig. \ref{fig:datapipe} .For real images, we collect an additional 6,000 samples from openly licensed news photography websites, including Reuters Pictures, Associated Press Images, BBC News In Pictures, UN Photo, and the International Committee of the Red Cross, ensuring consistent coverage of the same six domains.

\subsubsection{Everyday Life-Scenario Image Collection} 
We then gather 4,000 real-life images from COCO \cite{lin2014microsoft}, CC3M-val \cite{changpinyo2021conceptual}, and the publicly licensed photography platform Unsplash. To generate AI counterparts, we use captions from COCO \cite{lin2014microsoft} and CC3M-val \cite{changpinyo2021conceptual} as prompts for various generative models (SD 3.5 Large \cite{rombach2022high}, Flux.1 \cite{flux2024}, DALL-E3 \cite{ramesh2022hierarchical}, SDv1.4 \cite{rombach2022high}, SDv1.5 \cite{rombach2022high}, and Midjourney \cite{Midjourney2024}). To match the high-resolution photographic works, we also randomly selected a portion of AI-generated images from the Chameleon dataset \cite{yan2024sanity}.

Since real-world images include both high-resolution and low-resolution examples \cite{fan2024fake}, we also incorporate a portion of lower-resolution images generated by StyleGAN \cite{karras2019style} and ProGAN \cite{karras2017progressive} into RRDataset to better capture this characteristic. This step ensures that RRDataset more accurately reflects the real-world variations in image quality. Ultimately, we obtained a dataset of 4,000 images representing everyday scenes.

\subsection{Data transformation}
\label{sec33}
In this section, we describe how internet transmission and re-digitization are applied to our dataset. We also provide a detailed discussion in App. B on why these transformations are critical for AI-generated image detection and how they manifest in real-world scenarios.

\subsubsection{Internet Transmission}
Social media platforms typically compress images, causing reductions in resolution, compression artifacts, and detail loss. To evaluate detector robustness under realistic transmission conditions, we subjected all 10,000 real images and 10,000 AI-generated images from RRDataset to multiple rounds of sending through popular messaging and social platforms—Telegram, WeChat, Facebook, QQ, WhatsApp, X, Instagram, and Tinder. As summarized in Tab. \ref{tab:transmission}, each image underwent 2 to 6 transmission cycles, covering both cross-platform and single-platform settings. The resulting images were then added to the RRDataset test set, enabling a more faithful assessment of detector performance in real-world scenarios.

\begin{table}[ht]
    \centering
    \small 
    \renewcommand{\arraystretch}{1.0} 
    \setlength{\tabcolsep}{3pt} 
    \begin{tabular}{@{}cccc@{}} 
        \toprule
        Trans-Times & Percentage & Platforms & Self-Trans-Limit \\ 
        \midrule
        2 & 10\% & 1--2 & 2 \\ 
        3 & 25\% & 2--3 & 2 \\ 
        4 & 25\% & 2--4 & 3 \\ 
        5 & 25\% & 3--5 & 3 \\ 
        6 & 15\% & 4--6 & 3 \\ 
        \bottomrule
    \end{tabular}
    \caption{Cross-platform multi-transmission workflow, simulating three levels of information sharing: direct private messaging, limited-scale forwarding, and repeated forwarding of trending news.}
    \label{tab:transmission}
\end{table}

\subsubsection{Re-digitization}

Re-digitization refers to the process of converting a digital image into a physical format—such as printing or displaying it on a screen—and then converting it back into digital form, for instance, via scanning or photography. This process inevitably affects image quality, particularly in resolution and color fidelity, and may also introduce geometric distortions, adding another layer of complexity for detection.
We employ four common re-digitization methods:
\begin{itemize}
    \item Scanning a color printout.
    \item Photographing a color printout.
    \item Photographing an image displayed on a screen using various camera devices.
    \item Photographing a projected digital image with different camera devices.
\end{itemize}
Each method is applied with equal probability to 10,000 real images and 10,000 AI-generated images. The resulting re-digitized images—both real and AI-generated—are then incorporated into RRDataset.

\section{RRBench: Benchmarking Model Performance under Real-World Scenarios}

\subsection{Benchmark Setting}
\noindent \textbf{Detector Setup:} We evaluated 17 detection methods, including the latest state-of-the-art (SOTA) algorithms introduced at KDD 2025, AAAI 2025, and ICLR 2025. Specifically, our benchmarks incorporate CNNSpot \cite{wang2020cnn}, F3Net \cite{wei2020f3net}, GramNet \cite{liu2020global}, DIRE \cite{wang2023dire}, UnivFD \cite{ojha2023towards}, LNP \cite{liu2022detecting}, LGard \cite{tan2023learning}, AIDE \cite{yan2024sanity}, SSP \cite{chen2024single}, Fusing \cite{ju2022fusing}, Fredect \cite{frank2020leveraging}, DNF \cite{zhang2023diffusion}, NPR \cite{tan2024rethinking}, Freq-Net \cite{tan2024frequency}, SAFE \cite{li2024improving}, DRCT \cite{chen2024drct}, and C2P-clip \cite{tan2024c2p}.

\noindent \textbf{VLM Setup:} 
To further assess detection capabilities, we tested 10 contemporary vision-language models— gpt-4o-latest \cite{achiam2023gpt}, Claude-3-7-sonnet \cite{Anthropic2024}, Gemini-1-5 pro \cite{team2023gemini} , Gemini-2 flash \cite{team2023gemini} , GLM-4v-plus \cite{glm2024chatglm}, Grok-2-vision \cite{grok}, Qwen2.5-VL-72B \cite{bai2023qwen} , YI-vision \cite{young2024yi}, Moonshot-preview-vision-128k\cite{moonshot}, and Hunyuan-vision \cite{hunyuan}, using the prompt in App. D.2:








Each model was prompted to provide a JSON-based prediction—“AI-generated” or “Real”—based on its visual analysis. By including these vision-language systems, we aim to capture cutting-edge approaches to AI-generated image detection, enabling a broader comparison of performance across multiple methodological paradigms.

\noindent \textbf{Metric Setup:}  We use accuracy on real images and accuracy on AI-generated images as our primary evaluation metrics. Since the dataset is balanced across classes, recall and precision are straightforward to compute.

\noindent \textbf{Training Setup:} \noindent Following the setup outlined in Chameleon \cite{yan2024sanity} and GenImage \cite{zhu2024genimage}, we pretrain the detectors on GenImage-SD v1.4 and fine-tune them using RRDataset-subset. Training details and additional results are provided in App. C and App. D.

\begin{table*}[htbp]
\caption{Performance Comparison of 17 Detectors and 10 VLMs on RRDataset. All values are presented as percentages and represent the average results from three trials.  \textbf{``Fake"} denotes the accuracy on AI-generated images, while \textbf{``Real"} represents the accuracy on real images. Note: Since C2P-clip\cite{tan2024c2p} has not released its training code, we evaluate the model using its pre-trained weights.
}
\label{main_tab}
\centering
\begin{tabular}{c|cc|cc|cc|c}
\toprule
\multirow{2}{*}{Model} & \multicolumn{2}{c|}{Original} & \multicolumn{2}{c|}{Transmission} & \multicolumn{2}{c|}{Re-digitization} & Overall \\
 & Fake(\%) & Real(\%) & Fake(\%) & Real(\%) & Fake(\%) & Real(\%) & ACC(\%) \\
\toprule
\rowcolor{gray!20}
\multicolumn{8}{c}{\textbf{Detectors (Train on GenImage-SDv1.4 \& fine-tune on RRDataset)}} \\
DRCT-ConvB\cite{chen2024drct}
 & 93.52 & 95.52
 & 92.82(\textcolor{red}{-0.70}) & 95.09(\textcolor{red}{-0.43})
 & 64.34(\textcolor{red}{-29.18}) & 96.22(\textcolor{green}{+0.70})
 & 89.59 \\
DIRE\cite{wang2023dire}
 & 89.72 & 98.25
 & 90.34(\textcolor{green}{+0.62}) & 97.87(\textcolor{red}{-0.38})
 & 1.42(\textcolor{red}{-88.30}) & 98.89(\textcolor{green}{+0.64})
 & 79.42 \\
DNF\cite{zhang2023diffusion}
 & 90.62 & 99.05
 & 90.98(\textcolor{green}{+0.36}) & 94.45(\textcolor{red}{-4.60})
 & 0.05(\textcolor{red}{-90.57}) & 99.98(\textcolor{green}{+0.93})
 & 79.19 \\
AIDE\cite{yan2024sanity}
 & 78.95 & 78.94
 & 74.72(\textcolor{red}{-4.23}) & 78.75(\textcolor{red}{-0.19})
 & 76.04(\textcolor{red}{-2.91}) & 83.13(\textcolor{green}{+4.19})
 & 78.42 \\
GramNet\cite{liu2020global}
 & 81.34 & 74.65
 & 79.49(\textcolor{red}{-1.85}) & 75.69(\textcolor{green}{+1.04})
 & 79.45(\textcolor{red}{-1.89}) & 62.02(\textcolor{red}{-12.63})
 & 75.44 \\
CNNSpot\cite{wang2020cnn}
 & 72.42 & 89.09
 & 65.72(\textcolor{red}{-6.70}) & 88.78(\textcolor{red}{-0.31})
 & 43.12(\textcolor{red}{-29.30}) & 86.72(\textcolor{red}{-2.37})
 & 74.31 \\
LNP\cite{liu2022detecting}
 & 83.14 & 89.26
 & 38.23(\textcolor{red}{-44.91}) & 89.30(\textcolor{green}{+0.04})
 & 31.91(\textcolor{red}{-51.23}) & 91.05(\textcolor{green}{+1.79})
 & 70.48 \\
Fredect\cite{frank2020leveraging}
 & 79.60 & 75.93
 & 58.13(\textcolor{red}{-21.47}) & 82.11(\textcolor{green}{+6.18})
 & 46.34(\textcolor{red}{-33.26}) & 69.95(\textcolor{red}{-5.98})
 & 68.68 \\
NPR\cite{tan2024rethinking}
 & 49.21 & 96.18
 & 28.08(\textcolor{red}{-21.13}) & 97.13(\textcolor{green}{+0.95})
 & 38.65(\textcolor{red}{-10.56}) & 92.42(\textcolor{red}{-3.76})
 & 66.95 \\
Fusing\cite{ju2022fusing}
 & 87.24 & 92.46
 & 7.38(\textcolor{red}{-79.86}) & 99.04(\textcolor{green}{+6.58})
 & 30.79(\textcolor{red}{-56.45}) & 73.97(\textcolor{red}{-18.49})
 & 65.15 \\
SAFE\cite{li2024improving}
 & 98.29 & 88.22
 & 0.88(\textcolor{red}{-97.41}) & 98.85(\textcolor{green}{+10.63})
 & 2.29(\textcolor{red}{-96.00}) & 98.82(\textcolor{green}{+10.60})
 & 64.56 \\
Freq-Net\cite{tan2024frequency}
 & 76.08 & 82.18
 & 4.47(\textcolor{red}{-71.61}) & 98.64(\textcolor{green}{+16.46})
 & 37.48(\textcolor{red}{-38.60}) & 77.99(\textcolor{red}{-4.19})
 & 62.81 \\
F3Net\cite{wei2020f3net}
 & 65.82 & 71.35
 & 52.18(\textcolor{red}{-13.64}) & 75.49(\textcolor{green}{+4.14})
 & 31.16(\textcolor{red}{-34.66}) & 74.85(\textcolor{green}{+3.50})
 & 61.81 \\
UnivFD\cite{ojha2023towards}
 & 64.79 & 64.90
 & 44.61(\textcolor{red}{-20.18}) & 70.80(\textcolor{green}{+5.90})
 & 36.15(\textcolor{red}{-28.64}) & 75.69(\textcolor{green}{+10.79})
 & 59.49 \\
C2P-CLIP\cite{tan2024c2p}
 & 17.21 & 97.54
 & 28.82(\textcolor{green}{+11.61}) & 99.58(\textcolor{green}{+2.04})
 & 18.01(\textcolor{green}{+0.80}) & 90.29(\textcolor{red}{-7.25})
 & 58.58 \\
SSP\cite{chen2024single}
 & 61.29 & 64.54
 & 40.62(\textcolor{red}{-20.67}) & 70.33(\textcolor{green}{+5.79})
 & 32.58(\textcolor{red}{-28.71}) & 79.94(\textcolor{green}{+15.40})
 & 58.22 \\
LGrad\cite{tan2023learning}
 & 51.00 & 81.29
 & 18.86(\textcolor{red}{-32.14}) & 92.54(\textcolor{green}{+11.25})
 & 14.71(\textcolor{red}{-36.29}) & 88.27(\textcolor{green}{+6.98})
 & 57.78 \\
\toprule
\rowcolor{gray!20}
\multicolumn{8}{c}{\textbf{VLMs (Zero-shot)}} \\
GPT-4o-latest\cite{achiam2023gpt}
 & 96.30 & 92.68
 & 79.41(\textcolor{red}{-16.89}) & 90.01(\textcolor{red}{-2.67})
 & 69.23(\textcolor{red}{-27.07}) & 76.92(\textcolor{red}{-15.76})
 & 84.09 \\
Claude-3.7-sonnet\cite{Anthropic2024}
 & 85.12 & 94.57
 & 71.26(\textcolor{red}{-13.86}) & 96.17(\textcolor{green}{+1.60})
 & 62.34(\textcolor{red}{-22.78}) & 85.41(\textcolor{red}{-9.16})
 & 82.48 \\
Gemini-2-flash\cite{team2023gemini}
 & 72.10 & 98.43
 & 52.19(\textcolor{red}{-19.91}) & 97.41(\textcolor{red}{-1.02})
 & 46.11(\textcolor{red}{-25.99}) & 97.41(\textcolor{red}{-1.02})
 & 77.28 \\
Grok-2-vision\cite{grok}
 & 46.15 & 91.84
 & 52.12(\textcolor{green}{+5.97}) & 94.03(\textcolor{green}{+2.19})
 & 48.01(\textcolor{green}{+1.86}) & 81.63(\textcolor{red}{-10.21})
 & 68.96 \\
Gemini-1.5-pro\cite{team2023gemini}
 & 36.12 & 97.78
 & 36.36(\textcolor{green}{+0.24}) & 95.83(\textcolor{red}{-1.95})
 & 22.22(\textcolor{red}{-13.90}) & 88.64(\textcolor{red}{-9.14})
 & 62.83 \\
Qwen2vl-72B\cite{bai2023qwen}
 & 31.14 & 88.74
 & 22.87(\textcolor{red}{-8.27}) & 89.95(\textcolor{green}{+1.21})
 & 26.99(\textcolor{red}{-4.15}) & 92.55(\textcolor{green}{+3.81})
 & 58.71 \\
GLM4v-plus\cite{glm2024chatglm}
 & 22.16 & 90.57
 & 30.21(\textcolor{green}{+8.05}) & 86.01(\textcolor{red}{-4.56})
 & 38.16(\textcolor{green}{+16.00}) & 82.14(\textcolor{red}{-8.43})
 & 58.21 \\
Moonshot-vision\cite{moonshot}
 & 14.68 & 99.72
 & 16.24(\textcolor{green}{+1.56}) & 94.75(\textcolor{red}{-4.97})
 & 24.37(\textcolor{green}{+9.69}) & 96.84(\textcolor{red}{-2.88})
 & 57.77 \\
Hunyuan-vision\cite{hunyuan}
 & 28.57 & 91.84
 & 20.13(\textcolor{red}{-8.44}) & 86.06(\textcolor{red}{-5.78})
 & 32.14(\textcolor{green}{+3.57}) & 81.63(\textcolor{red}{-10.21})
 & 56.73 \\
YI-vision\cite{young2024yi}
 & 22.73 & 89.13
 & 20.45(\textcolor{red}{-2.28}) & 81.81(\textcolor{red}{-7.32})
 & 28.89(\textcolor{green}{+6.16}) & 78.26(\textcolor{red}{-10.87})
 & 53.55 \\

\bottomrule
\end{tabular}
\end{table*}

\subsection{Results and Analysis}
As shown in Tab. \ref{main_tab}, none of the 17 detection methods achieve saturated performance on RRDataset, with the best accuracy reaching only 89.59\%. This highlights both the ongoing complexity of AI-generated image detection and the crucial role of RRDataset in examining real-world challenges. Network transmission and re-digitization significantly degrade detection performance, underlining the dataset’s ability to reveal limitations overlooked by conventional evaluations. 

\noindent \textbf{Impact of Internet Transmission: } For internet-transmitted images, 14 detectors exhibit reduced fake accuracy, suggesting that compression artifacts, lower resolution, and color distortion lead to misclassification of AI-generated images as real. In particular, Freq-Net \cite{tan2024frequency}, Fusing \cite{ju2022fusing}, and SAFE \cite{li2024improving} experience accuracy drops of 71.61\%, 79.86\%, and 97.41\%, respectively, underscoring their lack of robustness. By contrast, DNF \cite{zhang2023diffusion} and DIRE \cite{wang2023dire}, which rely on diffusion-based denoising features, show only minor performance fluctuations under the same conditions. DRCT-ConvB \cite{chen2024drct}, which employs diffusion for image redrawing, similarly demonstrates strong resistance to transmission artifacts. Methods such as GramNet \cite{liu2020global} and AIDE \cite{yan2024sanity}, which focus on visual artifacts and noise patterns, also maintain relatively stable results when confronted with network-induced degradation.

\noindent \textbf{Impact of Re-digitization: } Re-digitization proves even more challenging: 16 of the 17 detection methods suffer decreases in fake accuracy. Notably, the diffusion-based DIRE \cite{wang2023dire} and DNF \cite{zhang2023diffusion} drop by 88.30\% and 90.57\%, respectively. In contrast, AIDE \cite{yan2024sanity} exhibits remarkable robustness, with only 1.89\% reduction in fake accuracy and a 4.19\% increase in real accuracy—likely owing to its reliance on CLIP-extracted semantic and contextual information, which remains largely intact after re-digitization.

\noindent \textbf{Vision-Language Models Performance: } Vision-language models demonstrate strong zero-shot classification capabilities, with GPT-4o \cite{achiam2023gpt} even surpassing 16 specialized detectors on the original data. However, their performance declines substantially under network transmission and re-digitization, indicating that reliance on internal model knowledge becomes a liability when images degrade in quality or exhibit color distortion.

\noindent Based on our findings, we can draw the following conclusions:
\begin{itemize}
    \item Diffusion-based methods are often robust to internet transmission, with models like DRCT-ConvB \cite{chen2024drct}, DNF \cite{zhang2023diffusion}, and DIRE \cite{wang2023dire} showing nearly unchanged performance against internet transmission.
    \item A hybrid strategy may contribute to the robustness of detection methods across different scenarios, as demonstrated by AIDE \cite{yan2024sanity}, which combines multiple expert models and uses hybrid features for detection, exhibiting the best robustness on RRBench, and by DRCT-ConvB \cite{chen2024drct}, which incorporates diffusion image reconstruction and contrastive training strategies, exhibiting the highest overall performance on RRBench.
    \item Vision-language models show considerable potential in AI-generated image detection, a promising yet overlooked aspect in existing detection research.
\end{itemize}

\section{Benchmarking Human Performance under Real-World Scenarios}
In this section, we conduct a comprehensive analysis of human performance on the RRDataset. Sec. \ref{51} introduces the evaluation system and process. Sec. \ref{53} presents a detailed analysis of human evaluation results.

\begin{figure*}[htbp]
  \centering
  \includegraphics[width=1.0\textwidth]{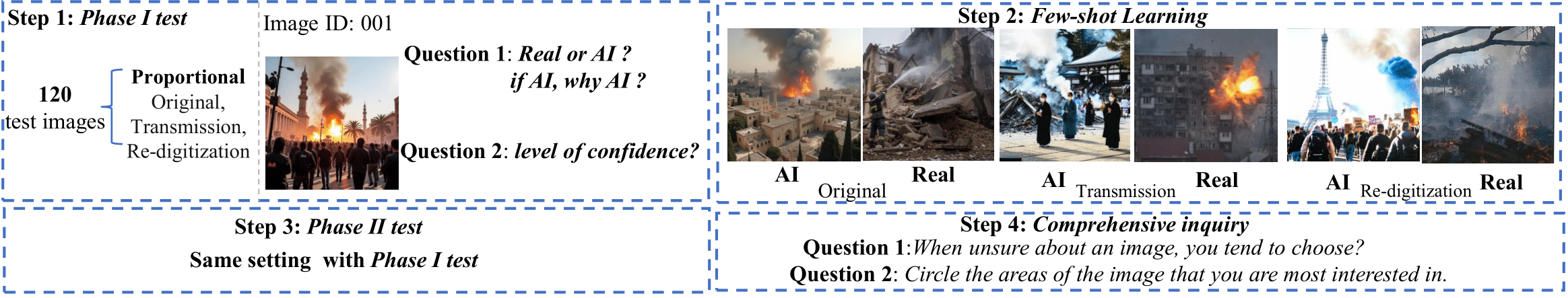}
  \caption{Human Benckmark Evaluation System.}
  \label{fig:bench}
\end{figure*}

\subsection{Human Benchmark Evaluation System}
\label{51}
Our human benchmark evaluation system consists of four steps, as shown in Fig. \ref{fig:bench}. A total of 192 participants were randomly assigned to either a special-scenario group or an everyday-scenario group. The special-scenario group viewed images from six high-impact categories: War \& Conflict, Disasters \& Accidents, Political \& Social Events, Medical \& Public Health, Culture \& Religion, and Labor \& Production.
To ensure reliability, there was no time limit, the test was conducted in a quiet environment on a standardized 4K display, and participants worked independently without any electronic aids or internet access.

\noindent \textbf{Phase One Testing.} To facilitate direct comparisons with RRBench, we selected 20 real and 20 AI-generated images from each of three categories—original, transmitted, and re-digitized—yielding 120 test images per participant. For each image, participants answered two questions: 

\noindent 1. \textit{Is this image AI-generated or real?}
If participants classified the image as AI-generated, they selected one of 14 reasons—five low-level criteria 
(\emph{texture}, \emph{edge}, \emph{clarity}, \emph{distortion}, \emph{overall hue}), 
five mid-level criteria (\emph{light \& shadow}, \emph{shape}, \emph{content deficiency}, 
\emph{symmetry}, \emph{reflection}), and four high-level visual criteria 
(\emph{layout}, \emph{perspective}, \emph{theme}, \emph{irreality}), 
consistent with FakeBench \cite{li2024fakebench}.

\noindent 2. \textit{How confident are you in this judgment?}
Responses were provided on a five-point Likert scale.

\noindent \textbf{Few-Shot Learning Phase.} Each participant viewed two additional images (original, transmitted, and re-digitized) drawn from RRDataset, ensuring no overlap with the main test set. 

\noindent \textbf{Phase Two Testing.} Phase Two replicated the Phase One procedure to evaluate whether 2-shot learning affected performance.

\noindent \textbf{Comprehensive Inquiry
.} Finally, two additional questions were posed:
1. \textit{When uncertain, do you tend to judge an image as AI-generated or real?} 2. \textit{Please highlight the area you focus on most.}
For the second question, participants were split into three subgroups of 32, each viewing six AI-generated images. Subgroup 1 viewed the original images, Subgroup 2 the transmitted versions, and Subgroup 3 the re-digitized versions. Each participant had 10 seconds to mark the region they found most telling, using a freeform drawing tool based on their immediate impression.

\subsection{Analysis of Human Evaluation Results}
\label{53}

\noindent \textbf{Overall Human Discrimination Ability: }
As shown in Tab. \ref{learning_performance}, the everyday life-scenario test group achieved an overall accuracy of 69.17\%, while the special-scenario test group reached 59.52\%, suggesting that humans are generally more adept at discerning AI-generated images in everyday contexts. For real images, the everyday life-scenario group achieved an average accuracy of 80.05\%, while the special-scenario group reached 79.40\%, indicating a negligible gap. However, for AI-generated images, the everyday life-scenario group attained 58.29\% accuracy, whereas the special-scenario group managed only 39.64\%. This difference indicates that in more sensitive, high-stakes contexts, humans are significantly less adept at identifying AI-generated images than they are in everyday life-scenarios. 
In both groups, internet transmission and re-digitization led to a significant, consistent drop in detection accuracy. During the first testing phase, these two factors reduced accuracy by 14.01\% and 14.29\%, respectively.

\begin{table}[htbp]
\caption{Human Benchmark Testing Accuracy Results. The numerical values in the table indicate accuracy (ACC). Values below 50\% are highlighted in red.}
\label{learning_performance}
\centering
\resizebox{\linewidth}{!}{%
\begin{tabular}{c|cc|cc|cc|c}
\toprule
\multirow{2}{*}{Group} & \multicolumn{2}{c|}{Original} & \multicolumn{2}{c|}{Trans.} & \multicolumn{2}{c|}{Re-digit.} & Overall \\
 & Fake & Real & Fake & Real & Fake & Real & ACC \\
\toprule
\rowcolor{gray!20}
\multicolumn{8}{c}{\textbf{Pre-learning}} \\
Everyday Life & 64.87 & 81.65 & \textcolor{red}{46.93} & 71.63 & \textcolor{red}{41.98} & 83.5 & 65.09 \\
Special-Scenario & \textcolor{red}{46.21} & 84.21 & \textcolor{red}{23.26} & 79.10 & \textcolor{red}{29.17} & 65.15 & 54.52 \\
\rowcolor{gray!20}
\multicolumn{8}{c}{\textbf{Post-learning}} \\
Everyday Life & 66.42 & 85.30 & 66.31 & 78.53 & 63.24 & 79.70 & 73.25 \\
Special-Scenario & 58.61 & 79.63 & \textcolor{red}{42.15} & 81.55 & \textcolor{red}{38.42} & 86.76 & 64.52 \\
\bottomrule
\end{tabular}%
}
\end{table}

\noindent \textbf{Analysis of Human Few-Shot Learning Ability: }  We compare the results of the first-stage and second-stage tests. Tab. \ref{learning_performance} highlights humans’ remarkable few-shot learning ability. For both the everyday-scenario group and the special-scenario group, accuracy for transmitted images increased by 13.14\% and 10.67\%, respectively, while accuracy for re-digitalized images rose by 8.73\% and 15.43\%. Overall, these improvements show that few-shot learning significantly boosts AI image recognition accuracy—from 42.07\% to 55.86\%, an increase of 13.79\%—and also raises accuracy for real images from 77.54\% to 81.92\%, a gain of 4.37\%.

\begin{figure}[htbp]
  \centering
  \includegraphics[width=0.4\textwidth]{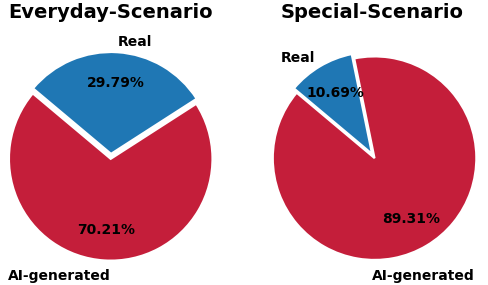}
  \caption{Trust Crisis Across Everyday Scenarios \& Special Scenarios.}
  \label{fig:trust}
\end{figure}

\begin{figure}[htbp]
  \centering
  \includegraphics[width=0.4\textwidth]{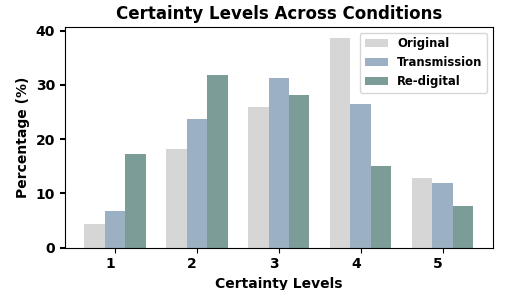}
  \caption{Uncertainty Levels across original images, transmission images and re-digital images where 1, 2, 3, 4, and 5 correspond to ``Very Uncertain," ``Somewhat Uncertain," ``Ambiguous," ``Somewhat Certain," and ``Very Certain," respectively.}
  \label{fig:confidence}
\end{figure}

\begin{figure*}[htbp]
  \centering
  \includegraphics[width=0.9\textwidth]{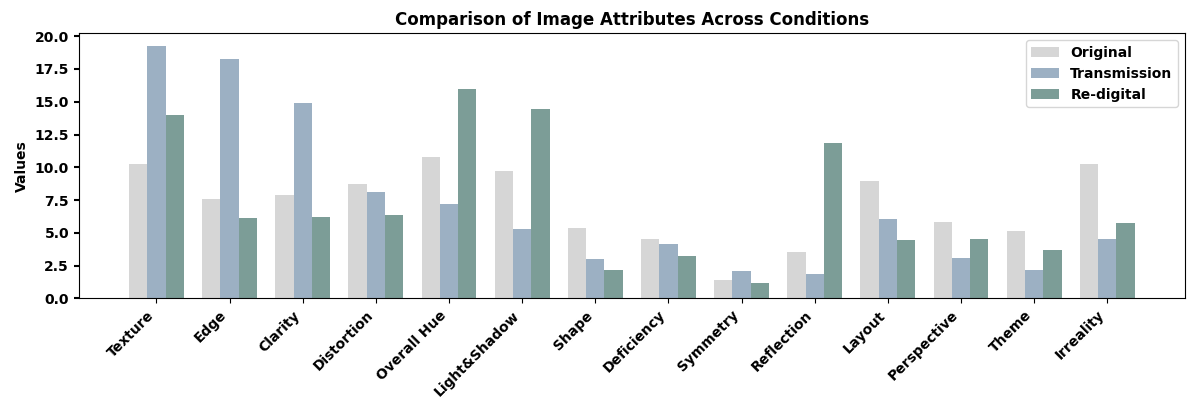}
  \caption{Comparison of Image Attributes Across Original, Transmission and Re-digital.}
  \label{fig:arr}
\end{figure*}

\noindent \textbf{Confidence Analysis: }For the five confidence levels provided—very uncertain, somewhat uncertain, ambiguous, somewhat certain, and very certain—we found that transmission and re-digitization significantly decreased participants’ confidence in their judgments as shown in Fig. \ref{fig:confidence}. For original images, the combined proportion of ``somewhat certain" and ``very certain" responses was 51.8\%, which dropped to 38.3\% for transmitted images and 22.7\% for re-digitized images. This indicates that transmission and re-digitization greatly impacted participants' confidence in their judgments.

\noindent \textbf{Trust Crisis Emerged in AIGC Era: }In the final part of the testing system, we asked all participants, “When you cannot decide whether an image is AI-generated or real, which type are you more inclined to choose?” Surprisingly, in Fig. \ref{fig:trust}, \textbf{70.21\%} of the everyday-scenario group assumed the image was AI-generated, while this figure rose to \textbf{89.31\%} in the special-scenario group. These findings indicate a significant crisis of trust in images’ origins, driven by the rapid advancement of AIGC technologies. The impact is especially severe for highly sensitive topics—such as War \& Conflict, Disasters \& Accidents, Political \& Social Events, Medical \& Public Health, Culture \& Religion, and Labor \& Production—where genuine news may be met with unwarranted skepticism, negatively influencing the broader information ecosystem. We therefore urge the community to develop more accurate, real-world-aligned detection methods to help mitigate this growing crisis of trust.

\noindent \textbf{Analysis of Judgemental Attribution: } As shown in Fig. \ref{fig:arr}, for the 14 possible reasons provided, transmission significantly increased the proportion of responses citing texture, edge, and clarity, rising from 25.78\% to 52.42\%. Re-digitization notably increased the proportion citing light\&shadow, reflection, and overall hue, from 24.08\% to 42.33\%. This suggests that the effects of transmission and re-digitization altered participants’ judgment criteria: transmission impacts are more related to reduced image quality, such as compression artifacts and detail loss, while re-digitization effects are more associated with color changes and loss of light and shadow details.

\subsection{Human-Inspired In-Context Learning Approch for VLM Detection}
Inspired by humans' remarkable few-shot learning abilities—where participants rapidly improved their robustness against network transmission and re-digitization with only a small number of samples—as well as the factors influencing human decision-making under these conditions, we aim to harness the strong zero-shot capability of VLMs to further enhance detection resilience.

Our in-context learning approach explicitly emphasizes the impact of transmission and re-digitization artifacts, guiding the VLM to focus on core image content while disregarding irrelevant distortions. Additional comparisons with other in-context learning strategies, as well as our method, are presented in App.D.

\begin{table}[h]
\centering
\caption{Performance Comparison for Human-Inspired Robustness-Oriented In-Context Learning. }
\label{gptandhumanlearn}
\scalebox{0.8}{
\begin{tabular}{lccc}
\toprule
 & \textbf{Original} & \textbf{Transmission} & \textbf{Redigital} \\
\rowcolor{gray!20}
\multicolumn{4}{c}{\textbf{Zero-shot}} \\
GPT-4o-latest & 94.49 & 84.71 & 73.08 \\
Claude-3.7-sonnet & 89.85 & 83.72 & 73.87 \\
Gemini-2-flash & 85.27 & 74.80 & 71.76 \\
Grok-2-vision & 68.99 & 73.08 & 64.82 \\
\rowcolor{gray!20}
\multicolumn{4}{c}{\textbf{Robustness-Oriented In-Context Learning}} \\
GPT-4o-latest & 95.67(\textcolor{green}{+1.18}) & 88.17(\textcolor{green}{+3.46}) & 78.58(\textcolor{green}{+5.50}) \\
Claude-3.7-sonnet & 92.26(\textcolor{green}{+2.41}) & 84.97(\textcolor{green}{+1.25}) & 77.76(\textcolor{green}{+3.79}) \\
Gemini-2-flash & 88.38(\textcolor{green}{+3.11}) & 74.99(\textcolor{green}{+0.19}) & 75.78(\textcolor{green}{+4.02}) \\
Grok-2-vision & 72.86(\textcolor{green}{+3.87}) & 71.52(\textcolor{red}{-1.56}) & 71.43(\textcolor{green}{+6.61}) \\
\bottomrule
\end{tabular}}
\end{table}

As shown in Tab. \ref{gptandhumanlearn}, our approach significantly enhances VLM robustness, particularly in re-digitization scenarios. Notably, on GPT-4o, our method achieves a \textbf{5.50\%} improvement in re-digitization robustness and a \textbf{3.90\%} increase in overall accuracy, reaching an average accuracy of \textbf{87.47\%}. This performance is approaching that of DCRT-ConvB (89.59\%), the strongest detector on RRBench to date. These findings underscore the potential of VLMs in AI-generated image detection.


\section{Conclusion}
In this paper, we introduced the RRDataset, rethinking the evaluation of AI-generated image detection from the perspective of real-world robustness. Our RRBench includes 17 detectors as well as comparisons with 10 VLMs, revealing a significant drop in accuracy for current detection methods under internet transmission and re-digitization conditions. Additionally, we developed the largest human benchmark to date, with 192 participants and 240 test images. We found that human accuracy dropped dramatically when faced with transmitted and re-digitized images; however, after few-shot learning, this effect of transmission and re-digitization was effectively mitigated. We hope this work encourages researchers to focus on the robustness of AI-generated image detection in real-world scenarios and to draw inspiration from humans' exceptional few-shot learning abilities in developing more robust and effective detection methods.

{
    \small
    \bibliographystyle{ieeenat_fullname}
    \bibliography{main}
}

\end{document}